\documentclass[runningheads]{llncs}

\usepackage[T1]{fontenc}
\usepackage{hyperref}
\usepackage{float}
\hypersetup{
    colorlinks=true,
    linkcolor=blue,
    filecolor=magenta,   
    citecolor=blue,
    urlcolor=cyan,
    pdftitle={EHRmonize},
    pdfpagemode=FullScreen,
    }
\usepackage{xcolor}
\usepackage{graphicx}
\usepackage{multirow}
\usepackage{booktabs}
\usepackage{tabularx}
\usepackage{array}
\usepackage{color, soul}

\begin{document}
\title{\texttt{EHRmonize}: A Framework for Medical Concept Abstraction from Electronic Health Records using Large Language Models}
\titlerunning{\texttt{EHRmonize}, a Framework for Medical Concept Abstraction}
%
\author{João Matos\inst{1} \and
Jack Gallifant\inst{2} \and
Jian Pei\inst{1} \and
A. Ian Wong\inst{1}\thanks{Corresponding Author: \email{a.ian.wong@duke.edu}}}
\authorrunning{Matos et al., 2024}
%
\institute{Duke University \and Massachusetts Institute of Technology}
\maketitle              
\begin{abstract}

Electronic health records (EHRs) contain vast amounts of complex data, but harmonizing and processing this information remains a challenging and costly task requiring significant clinical expertise. While large language models (LLMs) have shown promise in various healthcare applications, their potential for abstracting medical concepts from EHRs remains largely unexplored. We introduce \texttt{EHRmonize}, a framework leveraging LLMs to abstract medical concepts from EHR data. Our study uses medication data from two real-world EHR databases to evaluate five LLMs on two free-text extraction and six binary classification tasks across various prompting strategies. 
GPT-4o's with 10-shot prompting achieved the highest performance in all tasks, accompanied by Claude-3.5-Sonnet in a subset of tasks. GPT-4o achieved an accuracy of 97\% in identifying generic route names, 82\% for generic drug names, and 100\% in performing binary classification of antibiotics. While \texttt{EHRmonize} significantly enhances efficiency, reducing annotation time by an estimated 60\%, we emphasize that clinician oversight remains essential. Our framework, available as a Python package,\footnote{Package on \href{https://PyPI.org/project/ehrmonize/}{PyPI}, Repository on \href{https://github.com/aiwonglab/ehrmonize}{GitHub}, and Documentation on \href{https://ehrmonize.readthedocs.io/en/latest/index.html}{ReadTheDocs}.} offers a promising tool to assist clinicians in EHR data abstraction, potentially accelerating healthcare research and improving data harmonization processes.

\keywords{Large Language Models \and Electronic Health Records \and Chart Abstraction of Medical Concepts \and \texttt{EHRmonize}}
\end{abstract}
\section{Introduction}

The development of machine learning models in healthcare critically depends on large-scale, high-quality data. Electronic Health Records (EHRs) offer a rich source of such data, encompassing structured information generated during routine clinical practice, including vital signs, laboratory values, and clinical interventions \cite{valencia2023validation}. However, the full potential of EHR data remains largely untapped due to significant challenges in data processing. A primary obstacle in leveraging EHR data is the substantial variability in recording practices, both between and within hospital systems \cite{lester2022comparing}. This variability manifests in several ways:
\begin{itemize}
    \item \textbf{Inconsistent Terminology:} The same medical concept may be recorded differently across institutions or even within a single hospital. For example, in medication data, "dextrose 5\%" is an intravenous fluid for volume expansion (and the same as "D5W" and the normalized RxNorm concept "glucose 50 mg/ml"), but is different from "D50" and "magnesium sulfate 1 g in d5w", which are medical therapies~\cite{RxNorm_Technical_Documentation}.
    \item \textbf{Local Coding Systems:} Many healthcare institutions use local coding systems, making it difficult to compare/aggregate data across different sources.
    \item \textbf{Evolving Standards:} As medical knowledge and practices evolve, so do the terminologies and coding systems used in EHRs, further complicating long-term data harmonization efforts.
\end{itemize}
Current approaches to addressing these challenges often rely on manual abstraction (i.e., cleaning, categorization, and/or summarization) of concepts and chart review \cite{vassar2013retrospective,yin2022comparing}. However, these methods are time-consuming, labor-intensive, and prone to errors \cite{byrne2013comparison,lan2015automating,Sauer2022}. Moreover, the expertise required for accurate data abstraction is not always available, limiting the accessibility of EHR data for many researchers and potentially hindering progress in healthcare research \cite{wang2019}. Although there is a growing body of literature, most papers fail to be reproducible as the underlying codebases are not always shared.~\cite{pmlr-v68-johnson17a}

Large Language Models (LLMs) have emerged as a promising technology with the potential to revolutionize various aspects of medicine \cite{liu_summary_2023,chen2023generative}. Their ability to understand and generate human-like text has shown promise in tasks such as note summarization, clinical decision support, and medical education \cite{thirunavukarasu_large_2023}. Furthermore, LLMs have demonstrated significant encoded medical knowledge, as evidenced by their performance on medical question-answering benchmarks~\cite{jin_medqa}.

Given these capabilities, \emph{we hypothesize that LLMs can significantly improve workflow efficiency in abstracting medical concepts from EHR data}. By automating the categorization and harmonization of EHR entries, LLMs could potentially address many of the challenges associated with EHR data processing, thereby lowering barriers to entry for researchers and enabling more widespread use of EHR data in healthcare research and analytics.

In this paper, we introduce \texttt{EHRmonize}, a novel framework that leverages the power of LLMs to automate the cleaning and categorization of medical concepts in EHR data. Our work makes the following key contributions:

\begin{itemize}
    \item \textbf{LLM-based EHR Data Harmonization}: We present a novel approach to using LLMs for abstracting medical concepts in EHR data, addressing the critical need for efficient, scalable data harmonization methods. 
    \item \textbf{Curated Dataset: }We provide a curated dataset of medication data from MIMIC-IV \cite{johnson_mimic-iv_2023} and eICU-CRD \cite{pollard_eicu_2018}, enabling reproducibility of our findings and facilitating further research in this domain. This labeled dataset is made publicly available\footnote{Dataset on \href{https://huggingface.co/datasets/AIWongLab/ehrmonize}{HuggingFace}.}.
    \item \textbf{Comprehensive Evaluation:} We conduct an extensive evaluation of five state-of-the-art LLMs across various prompting strategies, encompassing two free-text tasks and six binary classification tasks. This evaluation provides insights into the capabilities and limitations of different LLMs in EHR data processing tasks.
    \item \textbf{Open-Source Implementation:} We release \texttt{EHRmonize} as an open-source PyPI package, implementing the use cases explored in this study and providing customizable modules for further applications. This contribution aims to foster collaboration and accelerate progress in the field of EHR data science.
\end{itemize}
By developing tools that automate the categorization and harmonization of EHR entries, \texttt{EHRmonize} aims to address critical challenges in EHR data processing, lower barriers to entry for researchers, and ultimately enable more widespread and efficient use of EHR data in healthcare research and analytics. In the following sections, we discuss related work, detail our methodology, present our findings, and discuss this work's implications and future directions.

\section{Related Work}
The challenge of harmonizing and extracting meaningful information from EHRs has been addressed through various approaches over the years. Traditional methods have included rule-based systems using hard-coded queries for automated data abstraction \cite{valencia2023validation} and cascading architectures for complex classification tasks \cite{dai2020adverse}. While effective for specific use cases, these approaches often lack flexibility and require significant effort to maintain as medical terminologies evolve.

Natural Language Processing (NLP) techniques have been widely applied to unstructured EHR data, with Named Entity Recognition (NER) being a key focus. Ahmad et al.~\cite{ahmad2023review} and Durango et al.~\cite{durango2023named} provide comprehensive reviews of NER techniques applied to clinical text, highlighting successes in identifying medical concepts despite linguistic variability challenges. However, these approaches often struggle with the variability of medical terminology across different EHR systems and require extensive manual input or task-specific fine-tuning, limiting their scalability and generalizability.

Efforts to standardize medical concepts have led to the development of tools like RxNorm~\cite{Waters2023,RxNorm_Technical_Documentation}. While these tools have made significant strides in concept matching across vocabularies, they often require extensive manual review, limiting their scalability. LLMs have opened new avenues for processing extensive medical texts at unprecedented speeds. Liu et al.~\cite{liu_summary_2023} and Chen et al.~\cite{chen2023generative} discuss the broader potential of LLMs to revolutionize various aspects of healthcare, from clinical decision support to medical education.

However, the application of LLMs in healthcare is not without challenges. Recent studies have highlighted concerns regarding the faithfulness~\cite{han2024towards} and bias~\cite{chen2024crosscare} of LLMs in medical contexts. In the domain of medication information processing, Gallifant et al.~\cite{gallifant2024language} demonstrated high performance in matching drug brand and generic terms using various LLMs, with GPT-4 achieving near-perfect accuracy. Nevertheless, their work also revealed limitations in handling more complex aspects of medication nomenclature. 
These findings underscore the need for specialized tools to manage the intricacies of medical drug data, which are crucial for developing comprehensive frameworks for AI-enabled pharmacovigilance and data harmonization~\cite{gallifant2024navigating}.

\texttt{EHRmonize} addresses a critical gap in this landscape by focusing on abstraction rather than mere extraction or labeling. We leverage LLMs for automated EHR data harmonization, aiming to capture and standardize higher-level concepts across diverse EHR systems. This approach combines the flexibility of machine learning with the nuanced understanding of medical language demonstrated by LLMs, potentially offering a more scalable and adaptable solution to the challenges of EHR data harmonization.

\section{Methods}
\texttt{EHRmonize} facilitates EHR data harmonization, addressing multidisciplinary collaboration challenges between data scientists and clinicians (Figure \ref{fig:context}). It comprises two components: corpus generation (SQL-based extraction of relevant text/concepts from EHR databases) and LLM inference (conversion of raw input to standardized classes via few-shot prompting) (Figure \ref{fig1}).


\definecolor{darkblue}{RGB}{0, 0, 139}
\definecolor{darkred}{RGB}{139, 0, 0}

\begin{figure}[H]
\caption{Example workflow and challenges in multidisciplinary clinical data science.}
\centering
\fbox{
\begin{minipage}{35em}
\scriptsize
\textbf{Context:} A multidisciplinary \textcolor{darkblue}{clinical} \textcolor{darkred}{data science} team is working with EHRs. The \textcolor{darkblue}{clinicians} agreed to include the patients on antibiotics, but exclude patients on anticoagulants. It is necessary to abstract medication and route names as "antibiotics" or "anticoagulants".\\
\textbf{Current Workflow, by role:}\\
\textbf{1. \textcolor{darkred}{Data Scientist}:} Queries unique medication names from EHRs and sends them to \textcolor{darkblue}{clinician}. \\
\textbf{2. \textcolor{darkblue}{Clinician}:} Maps medications into predefined classes and returns to the \textcolor{darkred}{data scientist}. \\
\textbf{Challenge:} Step 2 may involve manual labeling of thousands of entries with the help of a \textcolor{darkblue}{clinical expert}, which not all teams have access to.\\
\textbf{User Story:} As a \textcolor{darkred}{data scientist} working with EHR data, I want to automatically abstract medical concepts \textbf{as a first pass}, so that collaboration with \textcolor{darkblue}{clinical experts} is \textbf{more efficient}.
\end{minipage}
\label{fig:context}
}
\end{figure}

\begin{figure}[ht]
\includegraphics[width=0.95\textwidth]{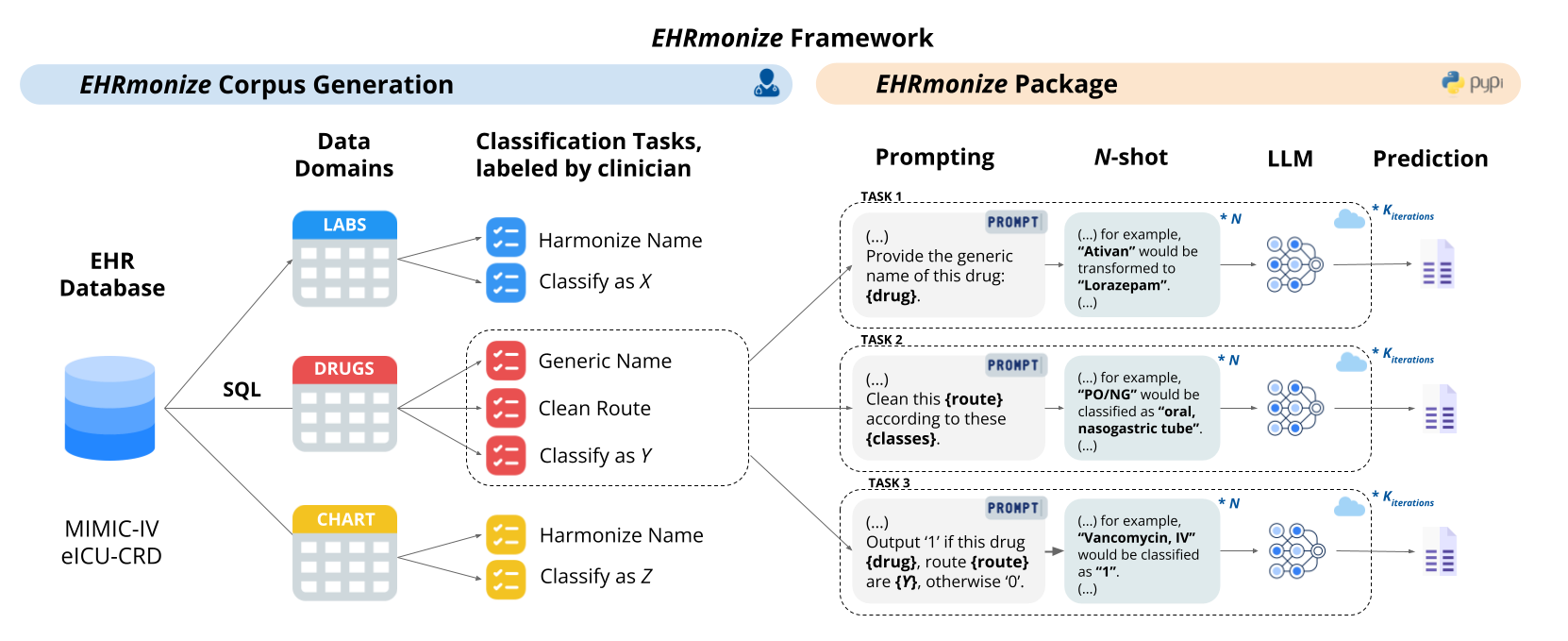}
\centering
\caption{Overall workflow of \texttt{EHRmonize}. Corpus generation from EHRs provides the data that needs categorization, across different domains and tasks, which is then fed to our package that employs LLMs to categorize the entries into predefined classes.}
\label{fig1}
\end{figure}

\textbf{Tasks:} We defined two task types: (1) free-text extraction of generic routes and drug names from raw entries, and (2) binary classification of (drug, route) pairs as antibiotic, anticoagulant, electrolytes, IV fluid, opioid analgesic, or stress ulcer prophylaxis. Data sources were MIMIC-IV \cite{johnson_mimic-iv_2023} and eICU-CRD \cite{pollard_eicu_2018}. Preprocessing involved SQL extraction of unique drug-route pairs, selection of top 200 prevalent entries per task, and manual labeling by a physician (AIW).

\textbf{Labeling:} \textit{Generic drug names:} Free-text drug names were translated to the lowercase generic name, matching either the clinical drug component, precise ingredient, or ingredient in RxNorm, a National Library of Medicine system to normalize medications~\cite{RxNorm_Technical_Documentation}. Salt names (e.g., hydromorphone hydrochloride to hydromorphone) were not included unless the active ingredient was shared across multiple salts (e.g., "metoprolol tartrate" vs. "metoprolol succinate"). Prescription strengths were not included (e.g., "hydromorphone hydrochloride 1mg" to "hydromorphone"). Concentrations were included for intravenous fluids and dextrose to disambiguate a precise drug (e.g., "normal saline 0.9\%" to "sodium chloride 9mg/ml"; "dextrose 50\%" to "glucose 500 mg/ml"). Medications with significant combinations (e.g., "pneumococcal 23-valent polysaccharide vaccine") kept all common RxNorm components but did not include valence. 
\textit{Generic routes:} Entries were transformed to the lowercase (no abbreviations) RxNorm classification (e.g., "IV" to "injectable product"; "PO/NG" to "oral product"). 
\textit{Binary classifications:} Six classes were one-hot encoded.

\textbf{Prompting:} Prompts included a specific task description, where we instruct the model to act like an experienced clinician, how the output format is expected to be, and how the expected class can be defined. When few-shot prompting was used, a few representative examples were provided. (Figure \ref{prompt_example})

\textbf{LLMs:} We assessed five models of 4 different families: Anthropic's Claude-3.5-Sonnet; Meta's Llama3-70B; Mistral's Mixtral-8x7B (via AWS's Bedrock API); and OpenAI's GPT-3.5-Turbo and GPT-4o. These models were selected due to their good performance in medical and non-medical benchmarks and cost-efficiency ratio.

\textbf{Experiments:} Besides the five different models, we explored different temperatures (0, 0.2, 0.5) and (0 to 10)-shot prompting. As the objective of \texttt{EHRmonize} is to improve efficiency in data cleaning, the time necessary to do manual annotation and \texttt{EHRmonize}'s output review was recorded and compared.

\begin{figure}[ht]
\caption{One-shot prompting example for the "IV fluid" binary classification task.}
\centering
\fbox{
\begin{minipage}{35em}
\scriptsize
You are a well trained clinician doing data cleaning and harmonization. You are given a raw drug name and administration route out of the EHR data below, within square brackets such as [drugname, route]. Please output \textbf{"1"} if [\textbf{"normal saline", "IV"}] is classified as \textbf{"IV fluid"}, otherwise \textbf{"0"}. "IV fluid" means "intravenous fluid given for the purpose of volume expansion". Consider the following example: An input drug name "sodium chloride 0.9\%" and route "IV" would be classified as "1". Please output nothing more than \textbf{"1"} or \textbf{"0"}. 
\end{minipage}
}
\label{prompt_example}
\end{figure}


\section{Results}

We labeled 398 entries from 14,604 and 8,803 unique medication-route pairs in eICU-CRD and MIMIC-IV databases, respectively (Table \ref{tab1}).

\begin{table}[ht]
    \caption{Characteristics of the medication entries in the labeled dataset.}
    \label{tab1}
    \centering
    \scriptsize
    \begin{tabularx}{\textwidth}{l c c c c c c c c c}
        \toprule
        \multirow{2}{*}{} & \multirow{2}{*}{} & \multicolumn{2}{c}{\begin{tabular}[c]{@{}c@{}}Free-Text \\ (\#Unique)\end{tabular}} & \multicolumn{6}{c}{\begin{tabular}[c]{@{}c@{}}Binary Tasks\\ (\#Positive)\end{tabular}} \\
        \cmidrule(lr){3-4} \cmidrule(lr){5-10}
        Database & N& \begin{tabular}[c]{@{}c@{}}Generic\\ Route\end{tabular} & \begin{tabular}[c]{@{}c@{}}Generic\\ Name\end{tabular} & Antibiotic & Anticoagulant & Electrolytes & \begin{tabular}[c]{@{}c@{}}IV\\ Fluids\end{tabular} & \begin{tabular}[c]{@{}c@{}}Opioid\\ Analgesic\end{tabular} & \begin{tabular}[c]{@{}c@{}}Stress ulcer\\ prophylaxis\end{tabular} \\
        \midrule
        MIMIC-IV & 198 & 6 & 83 & 8 & 13 & 17 & 22 & 12 & 8 \\
        eICU-CRD & 200 & 5 & 50 & 5 & 7 & 24 & 28 & 22 & 8  \\
        \bottomrule
    \end{tabularx}
\end{table}

\begin{figure}[ht]
\centering
\includegraphics[width=0.85\textwidth]{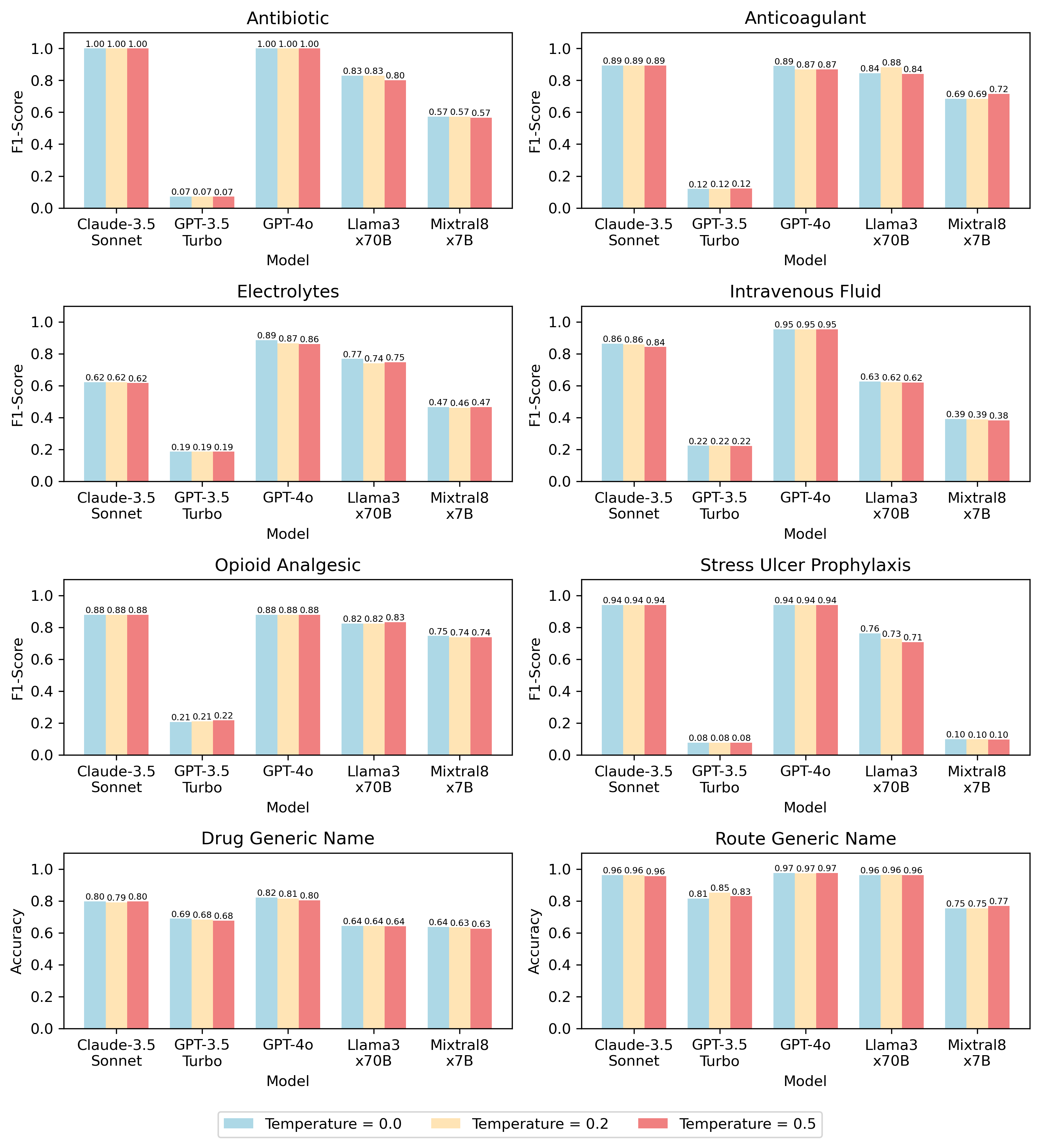}
\caption{LLM 10-shot performance across tasks and temperatures (398 samples).}\label{fig2}
\end{figure}

\textbf{Model Performance:} GPT-4o consistently outperformed other models, achieving an F1-score of 1.00 for antibiotic classification and 0.97 accuracy for route identification. Claude-3.5-Sonnet matched GPT-4o's performance in several binary classification tasks. GPT-3.5-Turbo, Llama3 70B, and Mixtral 8x7B showed lower performance (Figure \ref{fig2}). Generic drug name extraction proved challenging for all models, with GPT-4o achieving 0.82 accuracy.

\begin{figure}[ht]
\includegraphics[width=0.85\textwidth]{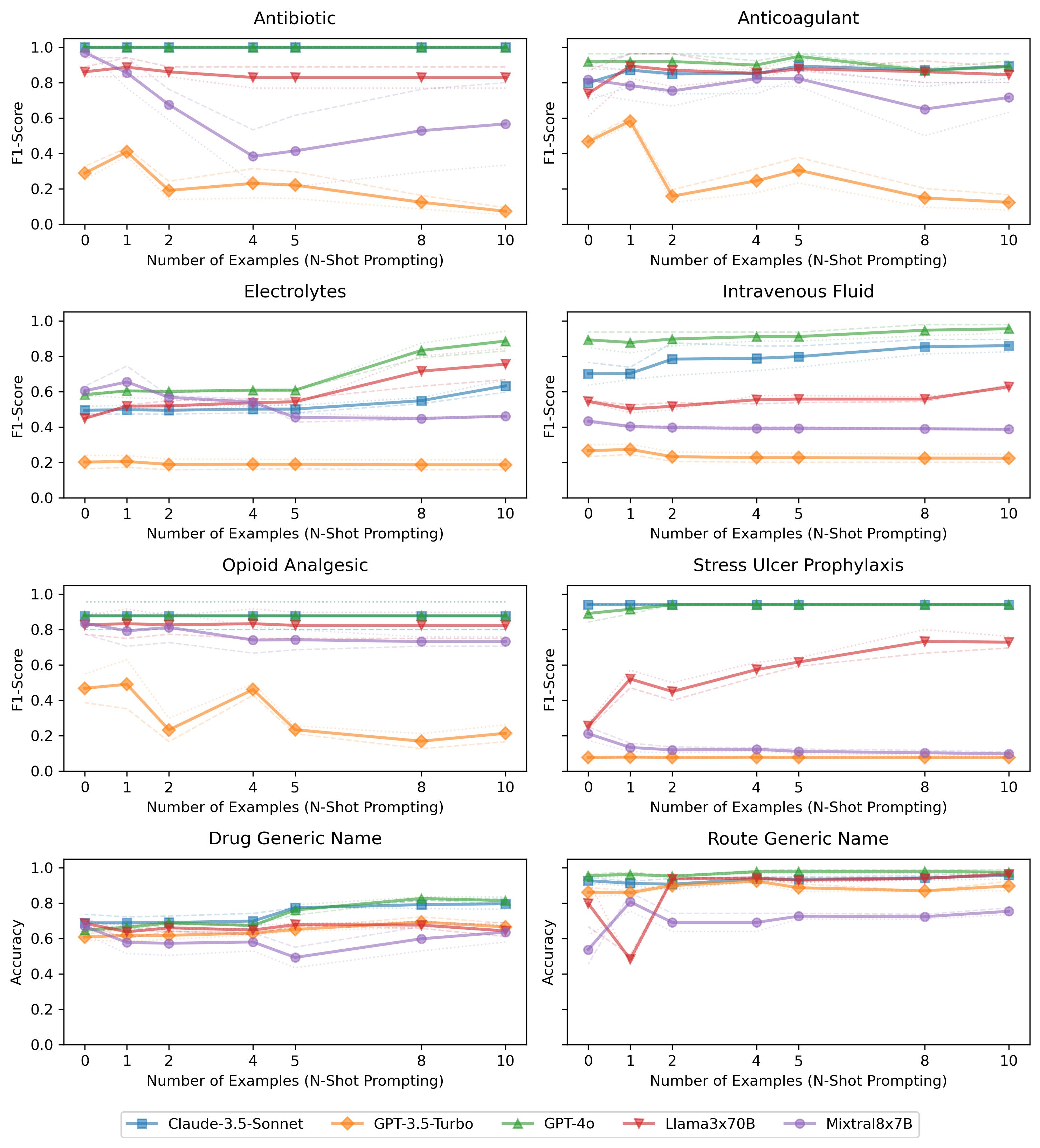}
\centering
\caption{LLM performance (temp. 0.2) with varying N-shot prompting across tasks.} \label{fig3}
\end{figure}

\textbf{N-shot Prompting:} GPT-4o and Claude-3.5-Sonnet exhibited stable, high performance with increasing examples. Unexpectedly, GPT-3.5-Turbo's performance declined as the number of examples increased, particularly in antibiotic, anticoagulant, and opioid analgesic tasks. Llama3 70B and Mixtral 8x7B maintained intermediate, relatively stable performance (Figure \ref{fig3}).

\textbf{Temperature Impact:} Variations in temperature up to 0.5 had minimal impact on model performance across tasks.

\textbf{Efficiency Gains:} \texttt{EHRmonize} significantly reduced annotation time, with savings of 67.9\% for MIMIC-IV and 60.4\% for eICU-CRD (Table \ref{tab2}).

\begin{table}[ht]
    \caption{Time (in minutes) spent in data annotation and in \texttt{EHRmonize} output review, using GPT-4o with 10-shot prompting.}
    \scriptsize
    \centering
    \begin{tabular}{@{}>{\raggedright}p{1.7cm}*{8}{>{\centering\arraybackslash}p{1.2cm}}@{}}
        \toprule
        Database & \multicolumn{4}{c}{MIMIC-IV} & \multicolumn{4}{c}{eICU-CRD} \\
        \cmidrule(lr){2-5} \cmidrule(lr){6-9}
        \parbox[t]{3cm}{\raggedright Task Type} & \parbox[t]{1.2cm}{\centering Generic} & \parbox[t]{1.2cm}{\centering Route} & \parbox[t]{1.2cm}{\centering Binary} & \textbf{Total} & \parbox[t]{1.2cm}{\centering Generic} & \parbox[t]{1.2cm}{\centering Route} & \parbox[t]{1.2cm}{\centering Binary} & \textbf{Total} \\
        \midrule
        Annotation & 6:03 & 3:56 & 5:53 & 15:52 & 4:46 & 3:40 & 8:31 & 16:57 \\
        Revision & 
        2:02 & 0:37 & 2:27 & 5:06 & 2:37 & 1:11 & 2:55 & 6:43 \\
        Corrections & 10/100 & 2/100 & 1/600 & 13/800 & 22/100 & 3/100 & 1/600 & 26/800\\
        Savings (\%) 
        & 66.4\% & 84.3\% & 58.4\% & \textbf{67.9\%} & 45.1\% & 67.7\% & 65.8\% & \textbf{60.4\%} \\
        \bottomrule
    \end{tabular}
    \label{tab2}
\end{table}


\section{Conclusion and Discussion}

\texttt{EHRmonize} demonstrates the potential of LLMs to abstract medical concepts from structured EHR data across multiple classification tasks. The framework demonstrated significant efficiency gains, reducing annotation time by approximately 60\%. This underscores \texttt{EHRmonize}'s potential to \textit{enhance}, rather than \textit{replace}, manual chart review by prepopulating options and allowing clinicians to focus on more complex abstraction tasks. 

Several limitations of this study warrant consideration. The dataset, while curated by a well-trained physician and supported by RxNorm materials, is limited in size (398 samples) and focused solely on medication data. Future research should aim to expand the dataset's volume and scope, incorporating other domains such as laboratory results and flowsheet data. Additionally, the current approach to N-shot example selection was deterministic; exploring the impact of example ordering could yield valuable insights into prompt engineering for medical NLP tasks.

Further avenues for improvement include incorporating semantic equivalence in free-text evaluation, implementing batching for enhanced efficiency, exploring retrieval-augmented generation (RAG) methods to extend N-shot examples, and investigating fine-tuning strategies for task-specific optimization. The potential of agentic approaches in managing abstraction workflows and ensuring consistency across outputs also merits exploration. Finally, regular evaluation on periodic data could facilitate the identification of concept drift, allowing the tool to adapt to evolving medical practices and terminologies.

\textbf{Prospect of application}: \texttt{EHRmonize}, now available as a Python package on PyPI, represents a significant step towards lowering barriers in EHR data research. Improving abstraction efficiency for structured data fields—a task often performed manually—has the potential to accelerate research and enable more comprehensive analyses of EHR data. 

\textbf{Disclosure of Interests:} AIW has received funding from NIMHD under U54MD012530. AIW has received support from AWS and CloudForce. All other authors have no competing interests.

\clearpage
\bibliographystyle{splncs04}
\bibliography{biblio}

\end{document}